\documentclass[runningheads]{llncs}
\usepackage[T1]{fontenc}
\usepackage{graphicx}

\usepackage{booktabs}
\usepackage{bm}
\usepackage{algorithm}
\usepackage{algorithmic}

\usepackage{amsmath}
\usepackage{amssymb}
\usepackage{amsfonts}

\begin{document}
\title{Label Selection Approach to \\ Learning from Crowds}
\iftrue
\author{Kosuke Yoshimura \and Hisashi Kashima}
\authorrunning{K. Yoshimura and H. Kashima}
\institute{Kyoto University, Kyoto, Japan\\
\email{yoshimura.kosuke.42e@st.kyoto-u.ac.jp \\ kashima@i.kyoto-u.ac.jp}
}
\else
\author{Anonymous Author(s)}
\authorrunning{Anonymous Author et al.}
\institute{Anonymous Institute}
\fi
\maketitle

\newcommand{\sn}{}

\begin{abstract}
Supervised learning, especially supervised deep learning, requires large amounts of labeled data. One approach to collect large amounts of labeled data is by using a crowdsourcing platform where numerous workers perform the annotation tasks. However, the annotation results often contain label noise, as the annotation skills vary depending on the crowd workers and their ability to complete the task correctly. Learning from Crowds is a framework which directly trains the models using noisy labeled data from crowd workers. In this study, we propose a novel Learning from Crowds model, inspired by SelectiveNet proposed for the selective prediction problem. The proposed method called Label Selection Layer trains a prediction model by automatically determining whether to use a worker’s label for training using a selector network. A major advantage of the proposed method is that it can be applied to almost all variants of supervised learning problems by simply adding a selector network and changing the objective function for existing models, without explicitly assuming a model of the noise in crowd annotations. The experimental results show that the performance of the proposed method is almost equivalent to or better than the Crowd Layer, which is one of the state-of-the-art methods for Deep Learning from Crowds, except for the regression problem case.
\end{abstract}
\section{Introduction}
Supervised learning, especially deep supervised learning, requires large amounts of labeled data.
One popular way to collect large amounts of labeled data is to use human annotators through crowdsourcing platforms human-intelligence tasks such as image classification~\cite{imagenet} and medical image classification~\cite{aggnet}.
To achieve an accurate prediction model, we require high-quality labeled data; however, since the annotation skills vary depending on crowd workers, collecting properly labeled data at the required scale has limitations.
Therefore, numerous studies were conducted for obtaining higher-quality labeled data based on the annotation results collected from crowd workers~\cite{generalcrowd,ds,confidence_score,multidim-wisdom,glad}.
A simple strategy is to collect responses from multiple crowd workers for each data instance, and to aggregate the answers using majority voting or averaging. 
However, the majority voting fails to consider natural human differences in the skills and abilities of crowd workers.
Dawid and Skene proposed a method to aggregate responses that considers workers' abilities~\cite{ds}, which was originally aimed at integrating doctors' examination results. 
Besides this, Whitehill et al. proposed GLAD, which further introduces the task’s difficulty~\cite{glad} into the model in addition to the workers’ abilities. 

A more direct approach is to train machine learning prediction models directly from crowdsourced labeled data rather than from high-quality labeled data, which is Learning from Crowds~\cite{lfc}. Raykar et al. proposed a method that alternates between parameter updating and EM algorithm-based answer integration within a single model using crowdsourced answers~\cite{lfc}. Additionally, they experimentally demonstrated that their proposed method performs better than a two-stage approach that first integrates crowdsourced labeled data by majority voting and then trains the model with this data as input~\cite{lfc}. 

In addition to classical machine learning methods, Learning from Crowds has also been studied for deep learning models~\cite{aggnet,adversarial_lfc,speelfc,crowdlayer}. Rodrigues and Pereira stated that the EM algorithm-based Deep Learning from Crowds method poses a new challenge: how to schedule the estimation of true labels by the EM algorithm for updates of learnable parameters~\cite{crowdlayer}. The reason is that the label estimation for the whole training dataset using the EM algorithm may be computationally expensive. Moreover, using the EM algorithm in each mini-batch may not contain sufficient information to correctly estimate the worker’s abilities. Rodrigues and Pereira proposed the Crowd Layer to solve this problem~\cite{crowdlayer}. 
However, a problem with Crowd Layer and other existing methods is that they require explicitly designed models of the annotation noise given by crowd annotators.
As the appropriate form of the noise model is task-dependent and is not necessarily obvious.

In this study, we propose a Label Selection Layer for Deep Learning from Crowds. The proposed method automatically selects training data from the crowdsourced annotations. It is inspired by SelectiveNet~\cite{SelectiveNet}, an architecture proposed in selective prediction. The Label Selection Layer can be added to any DNN model and can be applied simply by rewriting the loss function. The significant feature of the Label Selection Layer is that it does not require a generative model of worker annotation. This advantage is more pertinent in complex tasks such as structured output prediction rather than in simple tasks such as classification and regression problems, where annotation generation models are easier to represent in confusion matrices or additive noises. Another feature of the proposed method is that it does not use the EM algorithm. Therefore, the issue of using the EM algorithm stated in the previous study~\cite{crowdlayer} does not arise.

The contributions of this study are fourfold:
\begin{itemize}
    \item We propose a Label Selection Layer inspired by SelectiveNet as a novel Deep Learning from Crowds method.
    \item We proposed four variations of the Label Selection Layer: Simple, Class-wise, Target-wise, and Feature-based Label Selection Layer.
    \item We performed an experimental comparison of the proposed method with existing methods using real-world datasets to demonstrate the performance of the proposed method.
    \item We discussed the advantages and disadvantages of the proposed method based on the experimental results.
\end{itemize}

\section{Related Work}
This section summarizes related work from two aspects;
we first review the existing research related to Learning from Crowds, which learns prediction models from noisy crowdsourced annotations.
We next introduce selective prediction and related problems, which allow prediction models to neglect ``too hard'' instances.
In particular, we explain in detail SelectiveNet~\cite{SelectiveNet}, which is one of the typical solutions to the selective prediction problem, because this is the direct foundation of our our proposed method.

\subsection{Learning from crowdsourced annotations}
Extensive research was conducted for obtaining high-quality labels from noisy labels collected in crowdsourcing platforms~\cite{generalcrowd,complex_task,complex_matching,confidence_score,sequential_labeling,multidim-wisdom,glad}. As the quality of answers obtained from crowd workers is not always high, the most common method is to collect answers with redundancy~\cite{repeated_labels}. The simplest method is to collect responses from multiple workers for each task and merge these responses by majority voting. Dawid and Skene proposed a method to estimate the true label based on the estimated latent variables of each worker's ability using an EM algorithm~\cite{ds}. Their original goal was to estimate the correct medical examination result when there were multiple doctors' examinations for patients. Whitehill et al. proposed a label aggregation method by simultaneously modeling the worker's abilities and task difficulties as latent variables~\cite{glad}.
Welinder et al. proposed a response aggregation method that models worker skills and object attributes using multidimensional vectors~\cite{multidim-wisdom}.
Oyama et al. proposed a response aggregation method to collect the confidence scores of crowd workers for their own answers and to estimate both the confidence scores and the worker's ability~\cite{confidence_score}.
Baba and Kashima proposed a quality control method for more general tasks such as those requiring unstructured responses~\cite{generalcrowd}. Their approach is to have a worker's answers scored by another worker and to model respondent ability and scorer bias with a graded response model.
For complex tasks such as translation and ranking, Braylan and Lease proposed three modeling methods focusing on the distance between labels rather than the labels themselves. Their methods apply to tasks where requesters can appropriately define the distance function between labels~\cite{complex_task}.
Braylan and Lease also proposed a framework for response integration by decomposing and merging responses for several complex tasks, such as text span labeling and bounding box annotation~\cite{complex_matching}. This framework can handle multi-object annotation tasks that distance-based methods cannot.
As a specialized method for sequential labeling, Sabetpour et al. proposed AggSLC, which considers worker's responses, confidence, prediction results by machine learning model, and the characteristics of the target sequential labeling task~\cite{sequential_labeling}.
Wang and Dang proposed a method for generating appropriate sentences for a sentence-level task using a transformer-based generative model~\cite{wang2022generative}. This generative model inputs sentences that workers answered and outputs an integrated sentence. Wang and Dang also proposed a method to create pseudo-sentences as training data for the model~\cite{wang2022generative}. 

Extensive research has been conducted to obtain better-performing predictors using crowdsourced labeled data~\cite{aggnet,adversarial_lfc,speelfc,lfc,crowdlayer}. One intuitive approach is aggregating labels collected from multiple workers and then training them using the aggregated labels. By contrast, Raykar et al. proposed a framework called \textit{Learning from Crowds}, which directly trains a machine learning model end-to-end using the answers obtained from the crowd workers as input~\cite{lfc}. Their method alternately aggregates labels based on the EM algorithm and updates the model's parameters in a single model.
Kajino et al. stated that many learning-from-crowds approaches have non-convex optimization problems and proposed a new learning-from-crowds approach to make the problem convex by using individual classifiers~\cite{Kajino_2012}.
Chu et al. assumed that the causes of label noise can be decomposed into common worker factors and individual worker factors and proposed CoNAL, which models these factors~\cite{CoNAL}.
Takeoka et al. successfully improved the classifier's performance by using the information that workers answered that they did not know in the context of learning from crowds~\cite{unsure}.
Albarqouni et al. proposed a framework for estimating the true label from crowdsourced responses based on an EM algorithm and using it to train a deep learning model, in the context of a medical image classification problem~\cite{aggnet}.

Rodrigues and Pereira investigated the Learning from Crowds method based on the EM algorithm in case of deep learning models~\cite{crowdlayer}. The EM algorithm-based method had drawbacks such as it required adjustment of the timing of updating latent variables and learning parameters, and it was computationally expensive. Rodrigues and Pereira proposed \textit{Crowd Layer} that is directly connected to the output layer of the DNNs and acts as a layer that transforms the output results into the answers of each annotator~\cite{crowdlayer}. Although the nature of the Crowd Layer limits it learning to predict the labels correctly, their experiments have shown that DNN models with the Crowd Layer can produce a high-performance predictor.
Chen et al. proposed SpeeLFC, which can represent each annotator's answer in the form of conditional probability, given the prediction of the output layer, because of the lack of interpretability of the Crowd Layer parameters compared to the probabilistic approaches~\cite{speelfc}. 
They focused on the security aspect of learning-from-crowds systems. They stated that learning-from-crowds models are vulnerable to adversarial attacks. Therefore, they proposed A-LFC, which uses adversarial examples in the training step for this issue~\cite{adversarial_lfc}.

\subsection{Selective prediction} 
Machine learning models do not always return correct outputs. Therefore, selecting only highly accurate predictions is one way to obtain a more reliable predictor. This problem setup is called Selective Prediction~\cite{SelectiveNet}, Reject Option~\cite{Chow}, and Learning to defer~\cite{learning_to_defer}, with minor differences in definition among them. Research in selective prediction has been conducted since 1970, initially as machine learning with a reject option~\cite{Chow}. There is a wide range of research related to selective prediction. Here, we describe research on selective prediction in the case of deep learning models.

One simple and effective method is to set a threshold on the confidence score obtained from the prediction model~\cite{Chow}. Geifman and El-Yaniv determined the application of this framework to deep learning ~\cite{selective_classification_dnn}.
They also proposed SelectiveNet, which allows end-to-end selective prediction in a single model~\cite{SelectiveNet}. 

We explain the details of SelectiveNet~\cite{SelectiveNet}, which is the inspiration for our proposed method.
SelectiveNet is a neural network model for selective prediction, and it consists of a base model whose role is to extract features and three final output layers connected to it. The three final output layers are the prediction head, selective head, and auxiliary head. The prediction head predicts corresponding to the target variable~(denoted as $f^{\sn}(\textbf{x})$). The selective head selects whether to adopt the output of the prediction head~(denoted as $g^{\sn}(\textbf{x})$). The auxiliary head~(denoted as $h^{\sn}(\textbf{x})$) is a structure used to stabilize the training and, similar to the prediction head, it predicts corresponding to the target variable. However, SelectiveNet only uses the auxiliary head in the training step.

SelectiveNet aims to train the model so that for any given set $S=\{(\textbf{x}_n, y_n)\}_{n=1}^m$, where $\textbf{x}_n$ is the input feature for $n$-th instance, and $y_n$ is its true label, the empirical selection risk is minimal while keeping the empirical coverage at a predefined value. The empirical selective risk is defined by the following
\begin{equation}
    \hat{r}_\ell^{\sn}(f^{\sn}, g^{\sn}|S) = \frac{\frac{1}{m}\sum_{n=1}^{m}\ell(f^{\sn}(\textbf{x}_n), y_n)\cdot g^{\sn}(\textbf{x}_n)}{\hat{\phi}^{\sn}(g^{\sn}|S)},
\end{equation}
where $\ell(\cdot, \cdot)$ is any loss function. The empirical coverage is defined by the following
\begin{equation}
    \hat{\phi}^{\sn}(g^{\sn}|S) = \frac{1}{m}\sum_{n=1}^{m}g^{\sn}(\textbf{x}_n).
\end{equation}

Next, we show how SelectiveNet is trained. Given $c^{\sn}$ as the target coverage, the main loss function of SelectiveNet is as follows
\begin{equation}
    \mathcal{L}^{\sn}_{(f^{\sn}, g^{\sn})} = \hat{r}_\ell^{\sn}(f^{\sn}, g^{\sn}|S) + \lambda^{\sn} \Psi(c^{\sn} - \hat{\phi}^{\sn}(g^{\sn}|S)),
\end{equation}
where $\lambda^{\sn} > 0$ is a hyperparameter that balances the two terms, and $\Psi(a) = \max{(0, a)}^2$.
Here, the loss function for the auxiliary head is defined as
\begin{equation}
    \mathcal{L}^{\sn}_{h^{\sn}} = \frac{1}{m}\sum_{n=1}^{m}\ell(h^{\sn}(\textbf{x}_n), y_n),
\end{equation}
and the overall objective function of SelectiveNet is
\begin{equation}
    \mathcal{L}^{\sn} = \alpha^{\sn} \mathcal{L}^{\sn}_{(f^{\sn}, g^{\sn})} + (1 - \alpha^{\sn}) \mathcal{L}^{\sn}_{h^{\sn}},
\end{equation}
where $\alpha^{\sn}$ is a hyperparameter.

\section{Label Selection Layer}
\begin{figure*}[t]
  \centering
  \includegraphics[width=0.9\textwidth]{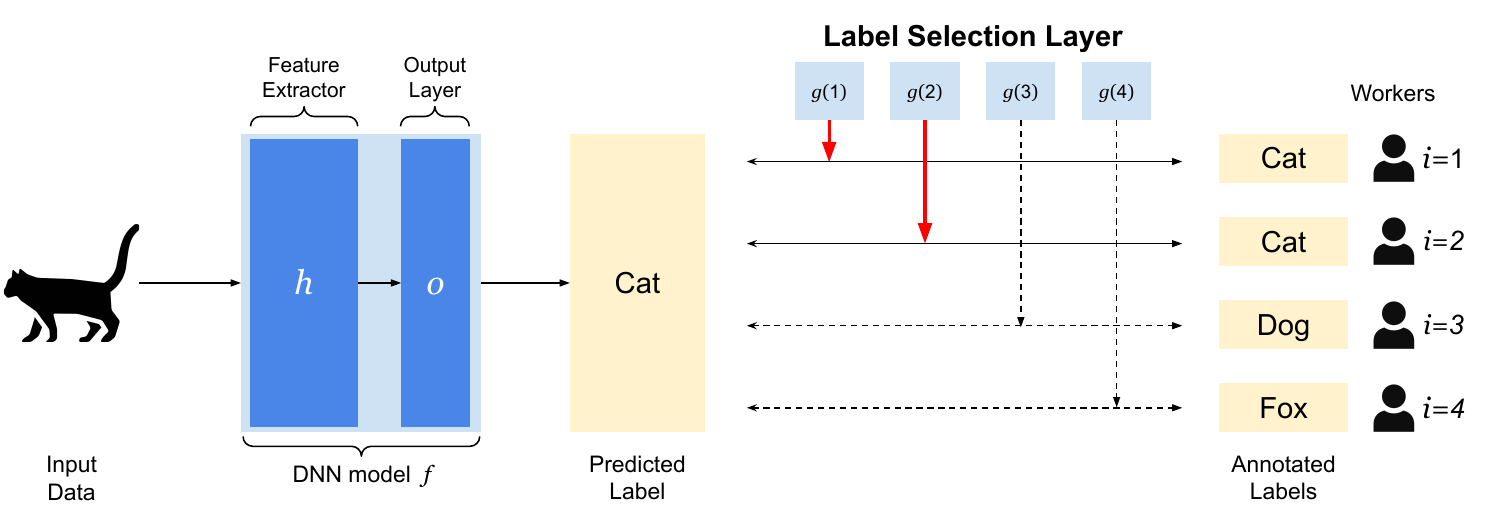}
  \caption{The structure using Simple Label Selection Layer in the case of four workers. The ground truth label of this input data is `Cat.' Because the annotated labels by workers $i=3, 4$ are incorrect, label selector $g(3)$ and $g(4)$ inactivate loss for labels by workers $i=3, 4$, `Dog' and `Fox,' respectively.}
  \label{proposedMethod}
\end{figure*}
We propose a novel Learning from Crowds method inspired by the SelectiveNet~\cite{SelectiveNet}, called \textit{Label Selection Layer}. The proposed method only selects reliable labels for training. It can be added to any DNN model and applied simply by rewriting the loss function.

\subsection{Problem Settings}
The goal of learning from crowds scenario is to train an accurate model using labels given by annotators. We define $f~(= o\circ h)$ as a DNN model, where $o$ is an output layer, and $h$ is a feature extractor. Let $\{(\textbf{x}_n, \textbf{y}_n)\}_{n=1}^{N}$ be a training data set, where $\textbf{x}_n \in \mathbb{R}^d$ is a $d$-dimension feature vector, and $\textbf{y}_n = \{y_n^i\}_{i\in\mathcal{I}}$ is an annotation vector provided by $|\mathcal{I}|$ annotators, with $y_n^i$ representing the label annotated to the $n$-th sample by the $i$-th annotator. Note that all annotators do not necessarily annotate all of the samples. Assume that each sample has a latent ground truth label.

In the $K$-class classification setting, for example, $y_n^i$ is an element of $\{1, 2, \ldots, K\}$, where each integer value denotes one of the candidate labels.

\subsection{Label Selection Layer}
The loss function of the proposed method reflects only the loss for the label selected by the Label Selection Layer. Let $g(i, l, \textbf{x}, h)$ be the label selector function corresponding to the label $l$ to a sample $\textbf{x}$ from the $i$-th worker using the feature extractor $h$ of the DNN model $f$.
In this case, we can write the adoption rate $\phi$ of the training data for all annotation set $A$ from crowd workers as
\begin{equation}
    \phi(g|A)=\frac{1}{|A|}\sum_n\sum_{y_n^i\in A_n}g(i, y_n^i, \textbf{x}_n, h),
\end{equation}
where $A_n$ means all annotation set for $n$-th sample.

From this equation, the empirical risk of a DNN model $f$ in learning to use annotated labels selectively is
\begin{equation}\label{empirical_risk}
    \hat{r}(f, g|A)=\frac{\frac{1}{|A|}\sum_{n}\sum_{y_n^i\in A_n}\ell(f(\textbf{x}_n), y_n^i)\cdot g(i, y_n^i, \textbf{x}_n, h)}{\phi(g|A)},
\end{equation}
where $\ell(\cdot, \cdot)$ is a loss function, which is defined depending on the target task. For example, we can use cross-entropy loss for the classification setting, mean square loss for the regression setting, and so on as $\ell(\cdot, \cdot)$. The numerator of Equation~(\ref{empirical_risk}) is the loss function value for each annotation label weighted by the value of the corresponding label selection layer.
To restrict the adoption rate $\phi$ to the training data using the hyperparameter $c$, the final loss function used in training can be expressed as
\begin{equation}\label{final_loss_function}
    \mathcal{L}(f, g)= \hat{r}(f, g|A) + \lambda \Psi(c - \phi(g|A)),
\end{equation}
where $\lambda>0$ is a hyperparameter that balances the two terms, and $\Psi(a)=(\max(0, a))^2$.

In the inference phase, we use only the predicted value $f(\textbf{x})$ as usual, and the label selection layer only selects useful training data in the training phase.

Here, we describe the differences between the proposed method and SelectiveNet. The most significant difference between the two is based on what is selected by the selection mechanism. The selective head $g^{\sn}$ of SelectiveNet is a weight for the loss $\ell(f^{\sn}(\textbf{x}_n), y_n)$ between the ground truth and the prediction result; therefore it takes a higher value for correct predictions. However, the label selection layer $g$ of the proposed method is a weight for the loss $\ell(f(\textbf{x}_n), y_n^i)$ between each worker's answer and the prediction. Therefore, it takes a higher value for workers' answers that are the same as (or close to) the prediction.

We discuss possible choices of the label selector function $g$. We propose four variations of the label selection layer: Simple, Class-wise, Target-wise, and Feature-based Label Selection Layer.

\subsubsection{Simple Label Selection Layer.}
Simple Label Selection Layer defines the label selector $g$ dependent only on each worker's one learnable weight scalar parameter $w_i$. Thus, we defined $g(i, l, \textbf{x}, h) = g(i) = \sigma(w_i)$, where $\sigma(\cdot)$ is a sigmoid function. Figure~\ref{proposedMethod} illustrates Simple Label Selection Layer. This approach assumes that each worker's annotation skills are constant for any given sample. Although the expressive power is not as significant owing to the small number of parameters, it has the advantage that it can be combined with any DNN model independent of the problem setup.

\subsubsection{Class-wise Label Selection Layer.}
Class-wise Label Selection Layer performs label selection in classification problems. It defined the label selector $g$ that is dependent on the learnable scalar weight parameters $w_{(i, l)}$ corresponding to each combination of worker and annotated label values. Thus, we defined $g(i, l, \textbf{x}, h)=g(i, l)=\sigma(w_{(i, l)})$. The method assumes that the annotation skills of each worker are dependent only on the label assigned by the worker and not on each sample. Although this method is used only for classification problems, it is more expressive than the Simple Label Selection Layer.

\subsubsection{Target-wise Label Selection Layer.}
Target-wise Label Selection Layer performs label selection in regression problems. It defined the label selector $g$ that is dependent on the learnable scalar weight parameters $w_i$ and scalar bias parameters $b_i$ corresponding to each worker, and target (continuous) variable $l$. Thus, we defined $g(i, l, \textbf{x}, h)=g(i, l)=\sigma((l \cdot w_{i} + b_{i})^{d_0})$, where $d_0$ is a hyperparameter. The method assumes that the annotation skills of each worker are dependent only on the continuous label assigned by the worker and not on each sample. Although this method is used only for regression problems, it is more expressive than the Simple Label Selection Layer.

\subsubsection{Feature-based Label Selection Layer.}
Feature-based Label Selection Layer defined the label selector $g$ that is dependent on the feature extractor output $h(\textbf{x})$ of the DNN model $f$ and learnable vector weight parameters $W_i$. Thus, we defined $g(i, l, \textbf{x}, h)=g(i, \textbf{x}, h)=\sigma(W_{i}^\top h(\textbf{x}))$. 
This method assumes that the correctness of each worker's annotation depends on who labels which sample.

\section{Experiments}
\begin{table}[t]
 \caption{The base model architecture used for the experiments with the LabelMe, the Movie Reviews, and CoNLL-2003 NER datasets. Descriptions of each layer and function are in the form of ``Input Size, Details.'' The number in parentheses indicates the kernel size.}
 \label{tb:base_model}
 \centering
  \begin{tabular}{llll}
   \cmidrule(r){1-2} \cmidrule(){3-4}
    \multicolumn{2}{c}{LabelMe} & \multicolumn{2}{c}{Movie Reviews} \\ \cmidrule(r){1-2} \cmidrule(){3-4}
    Flatten Layer & $4\times 4\times 512$  & Embedding & 1,000, 300 dim \\
    Flatten Layer & $4 \times 4 \times 512$  & Conv1d & 300 $\times$ 1,000, 128 Filters~(3) \\
    FC Layer & 8,192 & ReLU & 128 $\times$ 998 \\
    ReLU & 128 & MaxPool1d & 128 $\times$ 998, (5)\\
    Dropout & 128, 50\% & Dropout & 128 $\times$ 199, 50\% \\
    FC Layer & 128, 8units & Conv1d & 128 $\times$ 199, 128 Filters~(5) \\ \cmidrule(r){1-2}
     &  & MaxPool1d & 128 $\times$ 195, (5) \\ \cmidrule(r){1-2}
    \multicolumn{2}{c}{CoNLL-2003} & Flatten Layer & 128 $\times$ 39 \\ \cmidrule(r){1-2}
    Embedding & 109, 300 dim & FC Layer & 4992 \\
    Conv1d &  $300\times 109$, 512 Filters~(5) & ReLU & 32 \\
    Dropout & $109 \times 512$, 50\%  & FC Layer & 32, 1 unit \\ \cmidrule(r){3-4}
    GRU & $109\times 512$,  50 time steps & & \\
    FC Layer & 109, $50\times10$ units & & \\ \cmidrule(r){1-2}
  \end{tabular}
\end{table}
A comparative experiment between the proposed methods and existing methods is described. We conducted the experiments on three real datasets in classification, regression, and named entity recognition problems.
Table \ref{tb:base_model} shows the base model architectures used in each experimental setting. We used Adam~\cite{adam} to train all the models.
All the models used in the experiments were implemented in Pytorch~\cite{pytorch} and Pytorch-lightning~\cite{lightning}. The codes are available at \url{https://github.com/ssatsuki/label-selection-layer}.

\subsection{Image Classification}
\begin{table}[t]
 \caption{The results of accuracy comparison for the image classification dataset: LabelMe. Higher is better in all evaluation metrics. \textbf{Bold} indicates the best average value of the results of 100 trials, excluding those trained using Ground Truth Label.}
 \label{tb:accuracy}
 \centering
  \begin{tabular}{lcccc}
   \toprule
    Method & Acc. & OvR Macro AUC & Macro Prec. & Macro Rec.\\ \midrule
    w/ Ground Truth & .907~($\pm$.0056) & .994~($\pm$.0005) & .911 ~($\pm$.0053) & .911 ~($\pm$.0046)\\ \midrule
    \multicolumn{3}{l}{Crowd Layer~\cite{crowdlayer}} &  & \\
    \quad MW & .824~($\pm$.0266) & .979~($\pm$.0262) & .832 ~($\pm$.0419) & .833 ~($\pm$.0295) \\
    \quad VB & .824~($\pm$.0153) & .983~($\pm$.0105) & .834 ~($\pm$.0230) & .832 ~($\pm$.0157)\\
    \quad VW & .816~($\pm$.0096) & .983~($\pm$.0012) & .828 ~($\pm$.0079) & .825 ~($\pm$.0091)\\
    \quad VW$+$B & .824~($\pm$.0089) & .984~($\pm$.0010) & .834 ~($\pm$.0069) & .832 ~($\pm$.0083)\\ \midrule
    \multicolumn{3}{l}{(Proposed) Label Selection Layer} &  & \\
    \quad Simple & .796~($\pm$.0091) & .978~($\pm$.0012) & .811 ~($\pm$.0080) & .804 ~($\pm$.0091)\\ 
    \quad Class-wise & .796~($\pm$.0092) & .979~($\pm$.0012) & .813 ~($\pm$.0074) & .804 ~($\pm$.0094)\\ 
    \quad Feature-based & \textbf{.839~($\pm$.0114)} & \textbf{.985~($\pm$.0014)} & \textbf{.849 ~($\pm$.0096)} & \textbf{.848 ~($\pm$.0110)}\\ 
   \bottomrule
  \end{tabular}
\end{table}
We first address a multi-class classification problem by using the LabelMe dataset, which was also used in the previous study~\cite{crowdlayer}.
This dataset consists of 11,688 images and each image is assigned one of the eight candidate classes.
The dataset was divided into three parts: training, validation, and test data with 1,000, 500, and 1,180 labels, respectively. The validation and test data were assigned ground truth labels, and the validation data were used to determine hyperparameters and when to stop the training. Only labels annotated by 59 crowd workers were available as the training data, when training the proposed and conventional methods. The annotation results were collected from Amazon Mechanical Turk~\cite{orgLabelMe}.
We used the LabelMe data with the same preprocessing as in the previous work~\cite{crowdlayer}.

\begin{table*}[t]
 \caption{The results of comparison for the regression dataset: Movie Reviews. \textbf{Bold} indicates the best average value of the results of 100 trials, excluding those trained using Ground Truth Label.}
 \label{tb:error_movie_reviews}
 \centering
  \begin{tabular}{lccc}
   \toprule
    Method & MAE~($\downarrow$) & RMSE~($\downarrow$) & $R^2$~($\uparrow$)  \\ \midrule
    w/ Ground Truth & 1.061~($\pm$.050) & 1.346~($\pm$.066) & 0.447~($\pm$.055)\\ \midrule
    Crowd Layer~\cite{crowdlayer} & & & \\
    \quad S & 1.188~($\pm$.077) & 1.465~($\pm$.079) & 0.344~($\pm$.073) \\
    \quad B & \textbf{1.120~($\pm$.039)} & 1.465~($\pm$.047)& \textbf{0.393~($\pm$.041)} \\
    \quad S+B & 1.140~($\pm$.049) & \textbf{1.419~($\pm$.049)}& 0.386~($\pm$.043) \\ \midrule
    \multicolumn{2}{l}{(Proposed) Label Selection Layer} & & \\
    \quad Simple & 1.162~($\pm$.093) & 1.442~($\pm$.095) & 0.364~($\pm$.088) \\
    \quad Target-wise & 1.204~($\pm$.050) & 1.494~($\pm$.055) & 0.318~($\pm$.053) \\
    \quad Feature-based & 1.173~($\pm$.063) & 1.460~($\pm$.068) & 0.349~($\pm$.061) \\
   \bottomrule
  \end{tabular}
\end{table*}
We set the maximum number of epochs to 50, and the mini-batch size to 64. The evaluation metrics of this experiment are accuracy, one-vs-rest macro AUC, macro precision, and macro recall.

For comparison with the proposed methods, we used the base model trained using ground truth and the Crowd Layer combined with the output layer of the base model. As variations of the Crowd Layer, we selected MW, VW, VB, and VW+B proposed in \cite{crowdlayer}. MW, VW, VB, and VW+B use $\mathbf{W}^i\mathbf{\sigma}$, $\mathbf{w}^i\odot \mathbf{\sigma}$, $\mathbf{w}^i + \mathbf{b}$, and $\mathbf{w}^i\odot \mathbf{\sigma} + \mathbf{b}$ as annotator-specific functions of Crowd Layer respectively, where $\mathbf{W}^i$ is a $K\times K$ trainable weight matrix, $\mathbf{w}^i$ is a $K$ dimension trainable weight vector, $\mathbf{b}^i$ is a $K$ dimension trainable bias vector, and $\mathbf{\sigma}$ is a $K$ dimension output vector. We used the symbol $\odot$ as the Hadamard product operator.
About the hyperparameters of the proposed method, we fixed $\lambda=32$, and we selected $c$ with the best performance using the validation data.

We ran each method 100 times, and the average performance is shown in Table~\ref{tb:accuracy}. Comparison experiments using LabelMe dataset confirmed that one of the proposed methods, the Feature-based Label Selection Layer, performed better than the Crowd Layer in all evaluation metrics.
However, the performance of the remaining proposed methods is worse than that of the Crowd Layer.
Based on these results, it is possible that in the classification problem setting, the use of features for label selection can improve performance.

\subsection{Rating Regression}
We use the Movie Reviews dataset~\cite{movieReviews,labelMe} for rating regression. This dataset consists of English review comments collected from the Internet in \cite{movieReviews}, and the crowd workers on Amazon Mechanical Turk predicted the writers' review scores from the comments~\cite{labelMe}. Each annotator score takes a value between 0 and 10. However, they are not necessarily integer values. 
There were 5,006 reviews in English, divided into 1,498 and 3,508 review comments in training and test sets, respectively. All 135 workers predicted scores for this training set.
We normalized each score $y_n^i$ given by worker $i$ for review comments $n$ using the mean $\bar{y}$ and standard deviation $s$ of the ground truth score of the training data as
$\frac{y_n^i - \bar{y}}{s}$.
We converted each review comment by the tokenizer into a sequence of up to 1,000 tokens.

We used the pre-trained GloVe~\cite{GloVe} as an Embedding Layer and froze the embedding parameters. We set the maximum number of epochs to 100 and the mini-batch size to 128. The evaluation metrics of this experiment are mean absolute error~(MAE), root mean squared error~(RMSE), and $R^2$ score.

For comparison with the proposed method, we used a base model trained using ground truth and a Crowd Layer combined with the output layer of the base model. As variations of the Crowd Layer, we used S, B, and S+B proposed in \cite{crowdlayer}. S, B, and S+B use $s^i\mu$, $\mu + b^i$, and $s^i\mu + b^i$ respectively, where $s^i$ is a trainable scaling scalar, $b^i$ is a trainable bias scalar, and $\mu$ is an output scalar.

About the hyperparameters of the proposed methods, we set $\lambda = 32$ and selected the best $c$ from $\{0.1, 0.2, \ldots, 0.9\}$. The hyperparameter of the Target-wise Selective Layer is fixed $d_0=3$.

We ran each method 100 times, and the average performance is shown in Table \ref{tb:error_movie_reviews}. All of the proposed variants performed worse than the best Crowd Layer variant.
This is probably because the proposed method does not support learning with scaled or shifted annotation results, which is possible with the Crowd Layer.

\subsection{Named Entity Recognition}

\begin{table*}[t]
 \caption{The results of comparison for the named entity recognition dataset: CoNLL-2003 NER. \textbf{Bold} indicates the best average value of the results of 100 trials, excluding those trained using Ground Truth Label.}
 \label{tb:f1_score_ner}
 \centering
  \begin{tabular}{lccc}
   \toprule
    Method & Precision~($\uparrow$) & Recall~($\uparrow$) & $F_1$ score~($\uparrow$)  \\ \midrule
    w/ Ground Truth & 65.04~($\pm$2.24) & 66.33~($\pm$1.08) & 65.66~($\pm$1.35)\\ \midrule
    Crowd Layer~\cite{crowdlayer} & & & \\
    \quad MW~(pretrained w/ MV) & 54.95~($\pm$2.14) & \textbf{47.79~($\pm$1.61)} & \textbf{51.11~($\pm$1.61)} \\
    \quad VB~(pretrained w/ MV) & 61.33~($\pm$1.85) & 33.71~($\pm$2.07) & 43.47~($\pm$1.84)\\
    \quad VW~(pretrained w/ MV) & 62.42~($\pm$2.13) & 36.81~($\pm$1.75) & 46.28~($\pm$1.70)\\
    \quad VW+B~(pretrained w/ MV) & 61.31~($\pm$2.08) & 36.32~($\pm$1.63) & 45.59~($\pm$1.50)\\ \midrule
    (Proposed) Label Selection Layer & & & \\
    \quad Simple~(pretrained w/ MV) & \textbf{64.03~($\pm$1.53)} & 33.59~($\pm$1.88) & 44.03~($\pm$1.78)\\
    \quad Class-wise~(pretrained w/ MV) & 63.84~($\pm$1.41) & 35.33~($\pm$1.84) & 45.46~($\pm$1.71)\\
    \quad Feature-based~(pretrained w/ MV) & 63.49~($\pm$1.71) & 31.60~($\pm$2.07) & 42.17~($\pm$2.03)\\
   \bottomrule
  \end{tabular}
\end{table*}

As a learning task with a more complex output form, we address named entity recognition~(NER). We used a dataset consisting of CoNLL 2003 shared tasks~\cite{conll2003} and newly collected annotations based on it in \cite{labelMe}. We call it the CoNLL-2003 NER dataset in this study.
The CoNLL-2003 NER dataset consists of 9,235 text sequences, split into training, validation, and test data with 4,788, 1,197, and 3,250 sequences, respectively.\footnote{Note that the experimental setup in the previous study~\cite{crowdlayer} differs from our split setup because they only splits the data into train and test.} 
For the sequences in the training data, annotation results are collected from 49 workers on the Amazon Mechanical Turk in \cite{conllAmt}. 
The validation data were used to determine hyperparameters and when to stop the training.

We used the pre-trained GloVe~\cite{GloVe} as an Embedding Layer. The parameters of this Embedding Layer are trainable. We set the maximum number of epochs to 30, and the mini-batch size to 64. The evaluation metrics of this experiment are recall, precision, and $F_1$ score. For comparison with the proposed method, we used the base model trained using ground truth and the Crowd Layer combined with the output layer of the base model. As variations of the Crowd Layer, we selected MW, VW, and VW+B, the same as in the LabelMe experiments. 
In the CoNLL-2003 NLP dataset, we did not successfully train the Crowd Layer model when using the worker annotation set. Therefore, referring the published experimental code of Crowd Layer~\cite{crowdlayer}, we first pre-trained the DNN model in 5 epochs using the results of integrating the annotator's answers with Majority Voting. We connected the Crowd Layer to the DNN model and additionally trained it with worker annotations. We also trained the proposed method in the same way as we did with CrowdLayer. About the hyperparameters of the proposed method, we fixed $\lambda=32$, and we selected $c$ with the best performance using the validation data.

We ran each method 100 times, and the average performance is shown in Table~\ref{tb:f1_score_ner}. The experimental results show that all the proposed methods performed better than any variations of Crowd Layer on precision.

A significant feature of the Label Selection Layer is that no generative model of (errors of) worker annotation is required.
This advantage is particularly demonstrated in rather complex tasks such as structured output prediction, just like this NER task, in contrast with the Crowd Layer that requires correct assumptions about the mutation from true label sequences to ones given by the workers.

\section{Analysis}
\begin{table*}[t]
 \caption{For each training dataset, the mean and standard deviation over 30 trials of Pearson's correlation coefficient between the metric score based on each worker's answers and the mean of the outputs of the selection layer. N/A means that the proposed method cannot apply to the target dataset.}
 \label{tb:analysis}
 \centering
  \begin{tabular}{lrrrr}
   \toprule
   Dataset~(Metric) & Simple & Class-wise & Target-wise & Feature-based \\ \midrule
   LabelMe~(Acc.)& $0.54~(\pm 0.04)$ & $0.34~(\pm 0.00)$ & \multicolumn{1}{c}{N/A} & $0.44~(\pm 0.05)$\\
   MovieReviw~(RMSE)& $-0.68~(\pm 0.11)$ & \multicolumn{1}{c}{N/A} & $-0.15~(\pm 0.05)$ & $-0.61~(\pm 0.10)$\\
   CoNLL-2003~(Acc.) & $0.19~(\pm 0.18)$ & $0.90~(\pm 0.00)$ & \multicolumn{1}{c}{N/A} & $-0.03~(\pm 0.15)$ \\
   \bottomrule
  \end{tabular}
\end{table*}

Finally, we analyzed how well the proposed methods capture the quality of each worker's response.

Table \ref{tb:analysis} shows the Pearson correlation coefficients between the mean label selection score over training examples corresponding to each worker and the actual worker’s accuracy or RMSE.
For the LabelMe and CoNLL 2003 data sets, the methods with stronger positive correlations are more likely to be able to capture the quality of workers' actual responses.
For CoNLL 2003, the result of Simple Label Selection Layer shows a slightly positive correlation, while the result of the Feature-based Label Selection Layer shows almost no correlation on average.
This suggests that the NER task has a complex structure, and therefore a simple model is more likely to learn well than a complex model.
The fact that the Class-wise Label Selection Layer shows a very high correlation may be due to the dataset's characteristics.
The correlations might be overestimated due to the imbalance of labels, with most tokens having an O tag which means it is not a named entity.

For MovieReview, the method with the stronger negative correlation is considered to be better at capturing the quality of the worker's actual responses. All of the proposed methods successfully capture the quality of the worker's actual responses, as they show negative correlations to the RMSE of the MovieReview dataset. Since the absolute value of the correlation coefficient is smaller Target-wise than for the other two methods, we considered that it does not capture the quality of each worker's response well.

\section{Conclusion}
We proposed a new approach to deep learning from crowds: the Label Selection Layer~(LSL). The proposed method is inspired by SelectiveNet~\cite{SelectiveNet}, which was proposed in the context of selective prediction, and trains models using selective labels given by the workers. We can apply the LSL to any deep learning model by simply adding and rewriting the loss function. This flexibility is possible because the LSL does not assume an internal generative model. As variations of the LSL, we proposed the Simple, the Class-wise, the Target-wise, and the Feature-based LSL.
In the regression setting, the proposed method was worse than the Crowd Layer, but in the classification and the named entity recognition setting, the proposed method performed as well as or better than the Crowd Layer.
In particular, the experiments in the named entity recognition extraction setting showed that all proposed method variations outperform any Crowd Layer variation on precision. We demonstrated that the proposed method could be easily applied to complex tasks such as structured output prediction and shows high precision.

In future work, we will confirm whether the proposed method shows a higher performance in the Learning from Crowds problem setting for various structured data, so that it can be applied to different problem settings without significant changes. Additionally, we have to measure the performance improvement with different selector variations as it is likely that a more fine-grained selection of labels is possible using the labels by workers and the features provided as input to the model.
\bibliographystyle{splncs04}
\bibliography{references}

\begin{thebibliography}{10}
\providecommand{\url}[1]{\texttt{#1}}
\providecommand{\urlprefix}{URL }
\providecommand{\doi}[1]{https://doi.org/#1}

\bibitem{aggnet}
Albarqouni, S., Baur, C., Achilles, F., Belagiannis, V., Demirci, S., Navab,
  N.: Aggnet: Deep learning from crowds for mitosis detection in breast cancer
  histology images. IEEE Transactions on Medical Imaging  (2016)

\bibitem{generalcrowd}
Baba, Y., Kashima, H.: Statistical quality estimation for general crowdsourcing
  tasks. In: Proceedings of the 19th ACM SIGKDD International Conference on
  Knowledge Discovery and Data Mining (2013),
  \url{https://doi.org/10.1145/2487575.2487600}

\bibitem{complex_task}
Braylan, A., Lease, M.: Modeling and aggregation of complex annotations via
  annotation distances. In: Proceedings of The Web Conference 2020 (2020)

\bibitem{complex_matching}
Braylan, A., Lease, M.: Aggregating complex annotations via merging and
  matching. In: Proceedings of the 27th ACM SIGKDD Conference on Knowledge
  Discovery and Data Mining (2021)

\bibitem{adversarial_lfc}
Chen, P., Sun, H., Yang, Y., Chen, Z.: Adversarial learning from crowds.
  Proceedings of the AAAI Conference on Artificial Intelligence  (2022)

\bibitem{speelfc}
Chen, Z., Wang, H., Sun, H., Chen, P., Han, T., Liu, X., Yang, J.: Structured
  probabilistic end-to-end learning from crowds. In: Proceedings of the 29th
  International Joint Conference on Artificial Intelligence (2020)

\bibitem{Chow}
Chow, C.: On optimum recognition error and reject tradeoff. IEEE Transactions
  on Information Theory  (1970)

\bibitem{CoNAL}
Chu, Z., Ma, J., Wang, H.: Learning from crowds by modeling common confusions.
  In: Proceedings of the AAAI Conference on Artificial Intelligence (2021)

\bibitem{ds}
Dawid, A.P., Skene, A.M.: Maximum likelihood estimation of observer error-rates
  using the {EM} algorithm. Journal of the Royal Statistical Society. Series C
  (Applied Statistics)  (1979)

\bibitem{imagenet}
Deng, J., Dong, W., Socher, R., Li, L.J., Li, K., Fei-Fei, L.: Imagenet: A
  large-scale hierarchical image database. In: IEEE Conference on Computer
  Vision and Pattern Recognition (2009)

\bibitem{lightning}
Falcon, W., {The PyTorch Lightning team}: {PyTorch Lightning} (2019)

\bibitem{selective_classification_dnn}
Geifman, Y., El-Yaniv, R.: Selective classification for deep neural networks.
  In: Proceedings of the 31st International Conference on Neural Information
  Processing Systems (2017)

\bibitem{SelectiveNet}
Geifman, Y., El-Yaniv, R.: {S}elective{N}et: A deep neural network with an
  integrated reject option. In: Proceedings of the 36th International
  Conference on Machine Learning (2019)

\bibitem{Kajino_2012}
Kajino, H., Tsuboi, Y., Kashima, H.: A convex formulation for learning from
  crowds. In: Proceedings of the AAAI Conference on Artificial Intelligence
  (2012)

\bibitem{adam}
Kingma, D.P., Ba, J.: Adam: {A} method for stochastic optimization. In: 3rd
  International Conference on Learning Representations (2015)

\bibitem{learning_to_defer}
Mozannar, H., Sontag, D.: Consistent estimators for learning to defer to an
  expert. In: Proceedings of the 37th International Conference on Machine
  Learning (2020)

\bibitem{confidence_score}
Oyama, S., Baba, Y., Sakurai, Y., Kashima, H.: Accurate integration of
  crowdsourced labels using workers' self-reported confidence scores. In:
  Proceedings of the Twenty-Third International Joint Conference on Artificial
  Intelligence (2013)

\bibitem{movieReviews}
Pang, B., Lee, L.: Seeing stars: Exploiting class relationships for sentiment
  categorization with respect to rating scales. In: Proceedings of the 43rd
  Annual Meeting of the Association for Computational Linguistics (2005)

\bibitem{pytorch}
Paszke, A., Gross, S., Massa, F., Lerer, A., Bradbury, J., Chanan, G., Killeen,
  T., Lin, Z., Gimelshein, N., Antiga, L., Desmaison, A., Kopf, A., Yang, E.,
  DeVito, Z., Raison, M., Tejani, A., Chilamkurthy, S., Steiner, B., Fang, L.,
  Bai, J., Chintala, S.: Pytorch: An imperative style, high-performance deep
  learning library. In: Advances in Neural Information Processing Systems 32
  (2019)

\bibitem{GloVe}
Pennington, J., Socher, R., Manning, C.: {GloVe}: Global vectors for word
  representation. In: Proceedings of the 2014 Conference on Empirical Methods
  in Natural Language Processing (2014)

\bibitem{lfc}
Raykar, V.C., Yu, S., Zhao, L.H., Valadez, G.H., Florin, C., Bogoni, L., Moy,
  L.: Learning from crowds. Journal of Machine Learning Research  (2010)

\bibitem{labelMe}
Rodrigues, F., Lourenço, M., Ribeiro, B., Pereira, F.C.: Learning supervised
  topic models for classification and regression from crowds. IEEE Transactions
  on Pattern Analysis and Machine Intelligence  (2017)

\bibitem{conllAmt}
Rodrigues, F., Pereira, F., Ribeiro, B.: Sequence labeling with multiple
  annotators. Machine Learning  (2014)

\bibitem{crowdlayer}
Rodrigues, F., Pereira, F.C.: Deep learning from crowds. In: Proceedings of the
  32nd {AAAI} Conference on Artificial Intelligence (2018)

\bibitem{orgLabelMe}
Russell, B.C., Torralba, A., Murphy, K.P., Freeman, W.T.: {LabelMe}: A database
  and web-based tool for image annotation. International Journal of Computer
  Vision  (2008)

\bibitem{sequential_labeling}
Sabetpour, N., Kulkarni, A., Xie, S., Li, Q.: Truth discovery in sequence
  labels from crowds. In: 2021 IEEE International Conference on Data Mining
  (2021)

\bibitem{conll2003}
Sang, T.K., F, E., De~Meulder, F.: Introduction to the {{C}o{NLL}-2003} shared
  task: {Language-Independent} named entity recognition. In: Proceedings of the
  Seventh Conference on Natural Language Learning at {HLT}-{NAACL} 2003 (2003)

\bibitem{repeated_labels}
Sheng, V.S., Provost, F., Ipeirotis, P.G.: Get another label? improving data
  quality and data mining using multiple, noisy labelers. In: Proceedings of
  the 14th ACM SIGKDD International Conference on Knowledge Discovery and Data
  Mining (2008)

\bibitem{unsure}
Takeoka, K., Dong, Y., Oyamada, M.: Learning with unsure responses. In:
  Proceedings of the 34th AAAI Conference on Artificial Intelligence (2020)

\bibitem{wang2022generative}
Wang, S., Dang, D.: A generative answer aggregation model for sentence-level
  crowdsourcing task. IEEE Transactions on Knowledge and Data Engineering
  (2022)

\bibitem{multidim-wisdom}
Welinder, P., Branson, S., Perona, P., Belongie, S.: The multidimensional
  wisdom of crowds. In: Lafferty, J., Williams, C., Shawe-Taylor, J., Zemel,
  R., Culotta, A. (eds.) Advances in Neural Information Processing Systems
  (2010)

\bibitem{glad}
Whitehill, J., Wu, T.F., Bergsma, J., Movellan, J., Ruvolo, P.: Whose vote
  should count more: Optimal integration of labels from labelers of unknown
  expertise. In: Advances in Neural Information Processing Systems (2009)

\end{thebibliography}
\end{document}